\pdfoutput=1 
\documentclass{article}
\usepackage{spconf,amsmath,graphicx,bm, amsfonts,subcaption, multirow}
\usepackage{xcolor}
\usepackage[capitalize]{cleveref}
\usepackage{marvosym}
\usepackage{cite}
\crefname{section}{Sec.}{Secs.}
\Crefname{section}{Section}{Sections}
\Crefname{table}{Table}{Tables}
\crefname{table}{Tab.}{Tabs.}

\usepackage{fancyhdr}

\def\L{{\cal L}}

\title{Object-Aware Self-supervised Multi-Label Learning}
\name{Xu Kaixin$^{1}$\qquad  Liu Liyang$^{3}$\qquad  Zhao Ziyuan$^{1,2}$\qquad  Zeng Zeng$^{1,2}$\textsuperscript{,\Letter}\qquad Bharadwaj Veeravalli$^{3}$}
\address{$^{1}$Institute of Infocomm Research, A*STAR, Singapore\\
$^{2}$ Artificial Intelligence, Analytics And Informatics (AI$^3$), A*STAR, Singapore\\
$^{3}$School of Electrical and Computer Engineering, National University of Singapore, Singapore}
\begin{document}
%
\maketitle

\thispagestyle{fancy}
\fancyhead{}
\lhead{}
\vspace{-0.5pt}
\lfoot{\footnotesize{Copyright 2022 IEEE. Published in 2022 IEEE International Conference on Image Processing (ICIP), scheduled for 16-19 October 2022 in Bordeaux, France. Personal use of this material is permitted. However, permission to reprint/republish this material for advertising or promotional purposes or for creating new collective works for resale or redistribution to servers or lists, or to reuse any copyrighted component of this work in other works, must be obtained from the IEEE. Contact: Manager, Copyrights and Permissions / IEEE Service Center / 445 Hoes Lane / P.O. Box 1331 / Piscataway, NJ 08855-1331, USA. Telephone: + Intl. 908-562-3966.}}
\cfoot{}
\rfoot{}
\begin{abstract}

Multi-label Learning on image data has been widely exploited with deep learning models. However, supervised training on deep CNN models often cannot discover sufficient discriminative features for classification. As a result, numerous self-supervision methods are proposed to learn more robust image representations. However, most self-supervised approaches focus on single-instance single-label data and fall short on more complex images with multiple objects. Therefore, we propose an Object-Aware Self-Supervision (OASS) method to obtain more fine-grained representations for multi-label learning, dynamically generating auxiliary tasks based on object locations.
Secondly, the robust representation learned by OASS can be leveraged to efficiently generate Class-Specific Instances (CSI) in a proposal-free fashion to better guide multi-label supervision signal transfer to instances. 
Extensive experiments on the VOC2012 dataset for multi-label classification demonstrate the effectiveness of the proposed method against the state-of-the-art counterparts.

\end{abstract}
\begin{keywords}
Multi-label Learning, Self-Supervised Learning, Multi-instance Learning
\end{keywords}
\section{Introduction}
\label{sec:intro}

DCNNs (Deep Convolutional Neural Networks) have recently gained great success on a variety of computer vision tasks~\cite{zhao2019bira,voulodimos2018deep,zhao2020sea}, and have shown preliminary achievements on different Multi-label classification benchmarks~\cite{everingham2010pascal,lin2014microsoft,chua2009nus} compared to traditional image processing counterparts~\cite{zhang2013review, xu2020multi}, thanks to the strong capability of discovering high-level imaging features. 
Despite the power of deep learning, multi-label learning~(MLL) poses a more challenging scenario, where an image can contain multiple objects, each associated with a distinct class/label. Recent MLL works harness self-supervised representation learning (SSL) to tackle this problem. 
However, most of them did not overlap the capability of SSL and the objectives of MLL well, which will be discussed below.

This paper aims to improve the MLL performance by enhancing the model capacity to recognize more detailed local contextual features of the visual instances. Many Self-supervised Learning works~\cite{10.1007/978-3-319-46466-4_5, zhang2017split, zhang2016colorful, puzzlecam, donahue2019large, chen2020simple, misra2020self,Zeng,zhao2022self} explored to learn strong representations of the images to facilitate various tasks. Our baseline method PuzzleCAM~\cite{puzzlecam} tackles this by splitting the image into 4 even patches along the horizontal and vertical centers of the image and narrowing the discrepancy between the representation learned from the patch-level feature image-level ones. This poses a \emph{self-supervised} constraint on the model to learn such local contextual information of the visual instance without knowing the global information.
\begin{figure}[t]
    \centering
    \includegraphics[width=.82\linewidth]{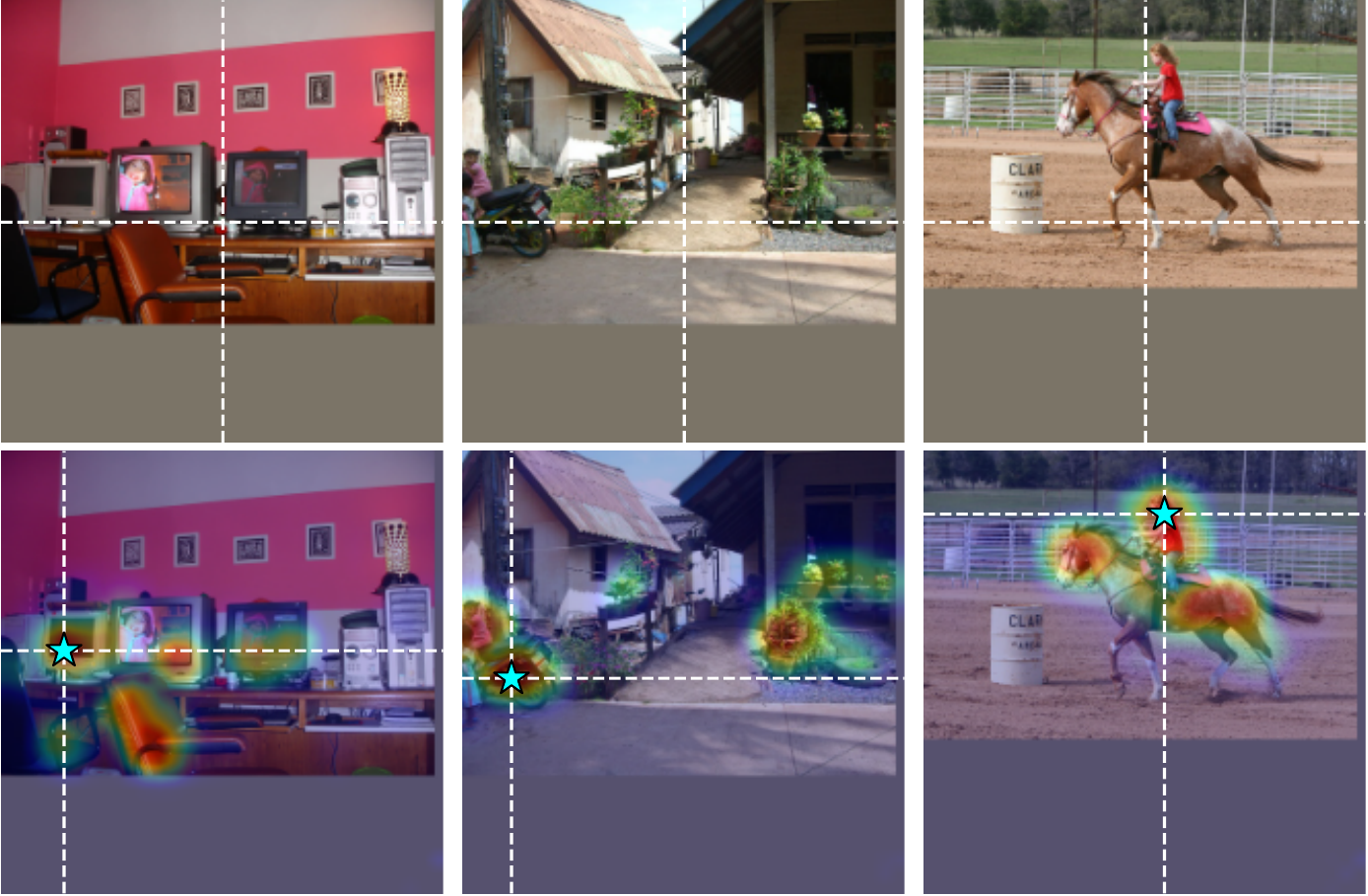}
    \caption{Illustration of the proposed patch generation strategy for Object-Aware Self-supervision. Note that the PuzzleCAM (first row) pads training images rather than resizes, undermining the quality of the patches. Our method (second row) is apparently more robust to multi-object scenario.}\vspace{-8pt}
    \label{fig:illu_cut}
\end{figure}

\begin{figure}
    \centering
    \includegraphics[width=.8\linewidth]{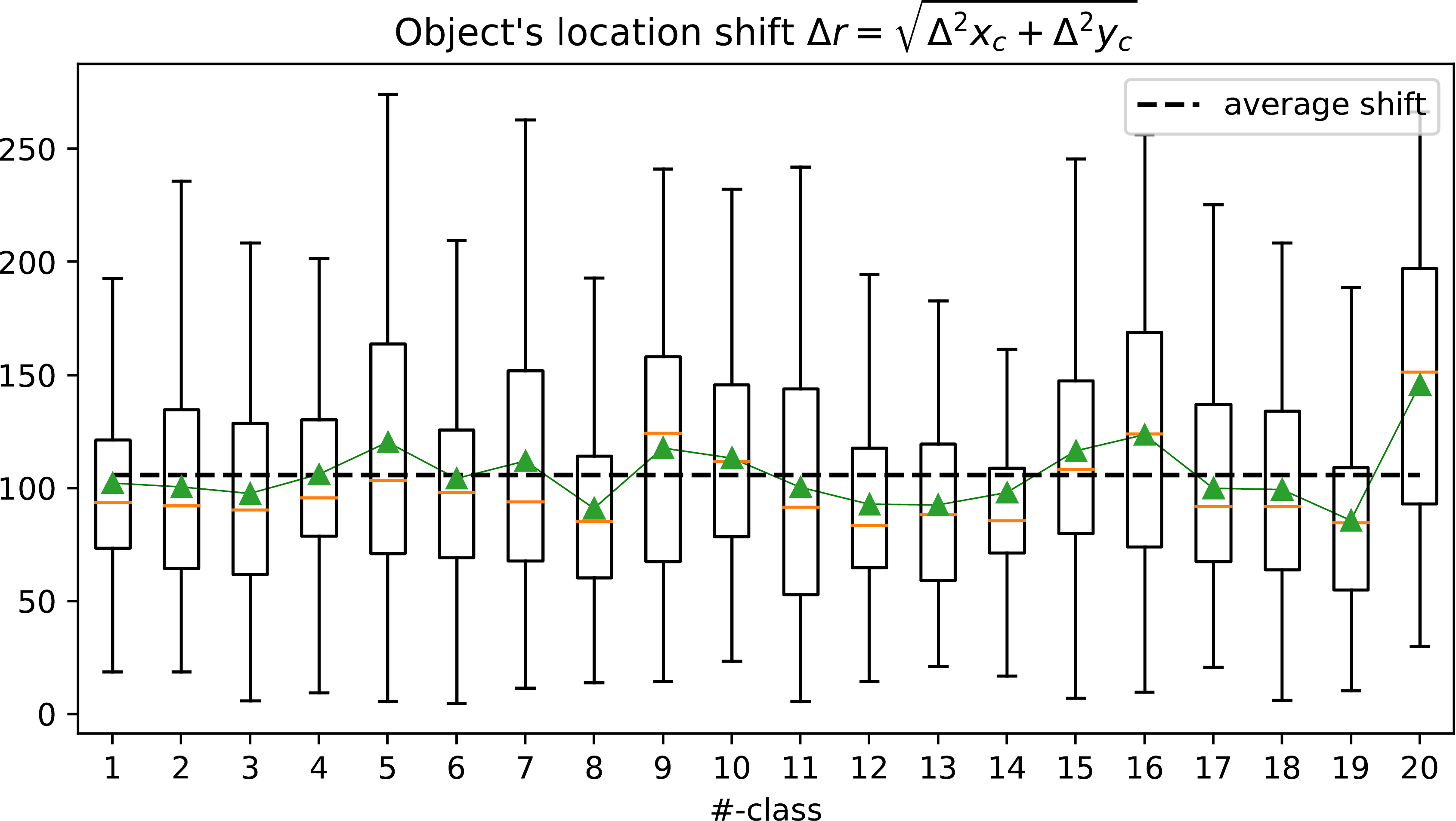}
    \caption{Object location shifts from image center. $\Delta x_c=\frac{x_l + x_h}{2} - \frac{W}{2}$ where $[x_l,x_h]$ is the horizontal range of the bounding box and $W$ is the wide of image. The statistics are obtained from VOC2012 \texttt{train} set. Object center clearly shifted from center by arround $100$ pixels on average with large variances on most classes, when image size is $512\times 512$. }
    \label{fig:stats_box}\vspace{-12pt}
\end{figure}

However, as illustrated in Fig.~\ref{fig:illu_cut}, one major drawback of this approach is that it would only impose sufficient supervision under the assumption that objects always fall in the center, \emph{i.e.}, single-object scenario. However, such assumption no longer stands in MLL, where instead the objects' location and scale are uncertain. According to our statistical analysis (\cref{fig:stats_box}), objects locations and scales become drastically unpredictable under the MLL setting. 
As a result, \cite{puzzlecam} may not provide sufficient supervision.
Therefore, we propose to generate tiled images more robustly under the MLL scenario and force the objects, i.e. visual instances be more evenly broken into each patch, so that the auxiliary supervision can effectively impose on the objects, \emph{i.e.,} Object-Aware Self-supervision (OASS). To further promote robust contextual knowledge learning, we adopt EMA~\cite{tarvainen2017mean} method in knowledge distillation for patch-level feature extraction.


Besides, since one label only corresponds to an/some instance(s) in the above setting, it is more ideal to explicitly optimize multi-label learner with instance-level representations, so that each semantic can focus on a fine-grained set of \emph{class-specific and instance-related} traits rather than being exposed to less decisive image-level features. Therefore we perform MLL at instance-level in opposed to previous MLL works~\cite{chen2019learning, wang2020multi}. Since the abovementioned OASS module provides robust object related visual cues from the image, we can efficiently generate such class-specific instance-level representations by leveraging class-wise attention maps on OASS features. Existing weakly-supervised object detection works~\cite{wang2020instance, bilen2016weakly, MCAR_TIP_2021} follow the similar idea, however, they require to manually extract objects from the image which is much less efficient. 
Experiments show the effectiveness of our method and its superiority against the baselines. 

We summarize our contributions in this paper as below:
\begin{itemize}
    \vspace{-6pt}
    \item We propose a new self-supervised constraint to serve the MLL task, \emph{i.e.}, Object-Aware Self-supervision (OASS) module to better learn local contextual representations under the multi-label scenarios.
    \vspace{-6pt}
    \item We generate Class-specific Instance~(CSI) from rich representations learned by the OASS and perform instance-level MLL efficiently.
    \vspace{-6pt}
    \item We conduct extensive experiments on multi-label classification dataset, demonstrating the advances in performance against various counterparts.
\vspace{-6pt}
\end{itemize}

\begin{figure*}[t]
    \centering
    \includegraphics[width=.75\textwidth]{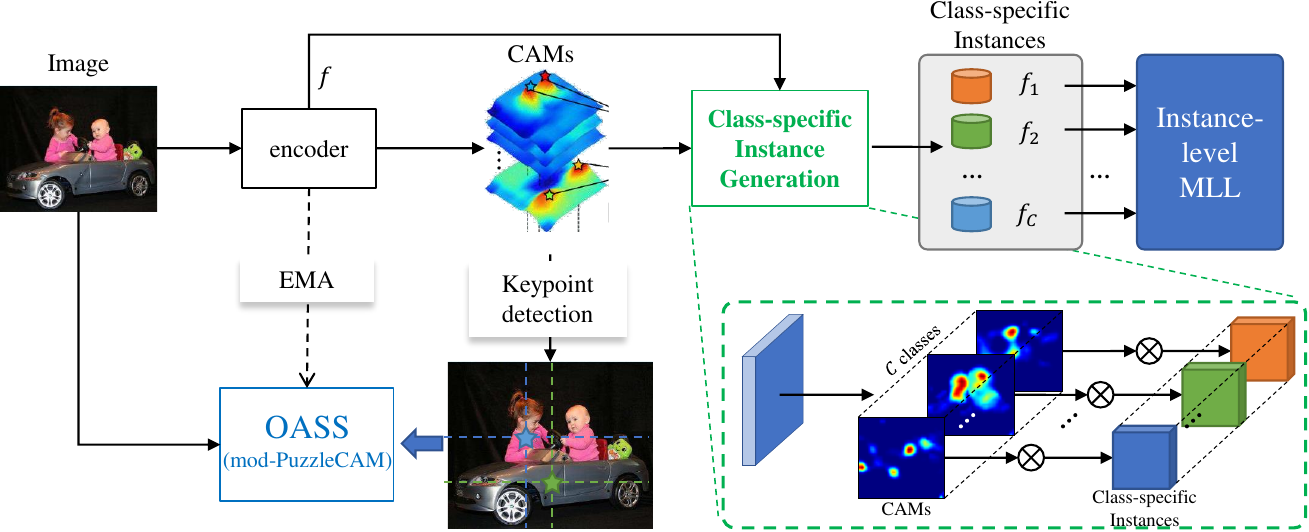}
    \caption{Overall Pipeline of our proposed method.}\vspace{-12pt}
    \label{fig:pipeline}
\end{figure*}

\begin{figure}[t]
    \centering
    \begin{subfigure}[b]{0.24\linewidth}
        \centering
         \includegraphics[width=\textwidth]{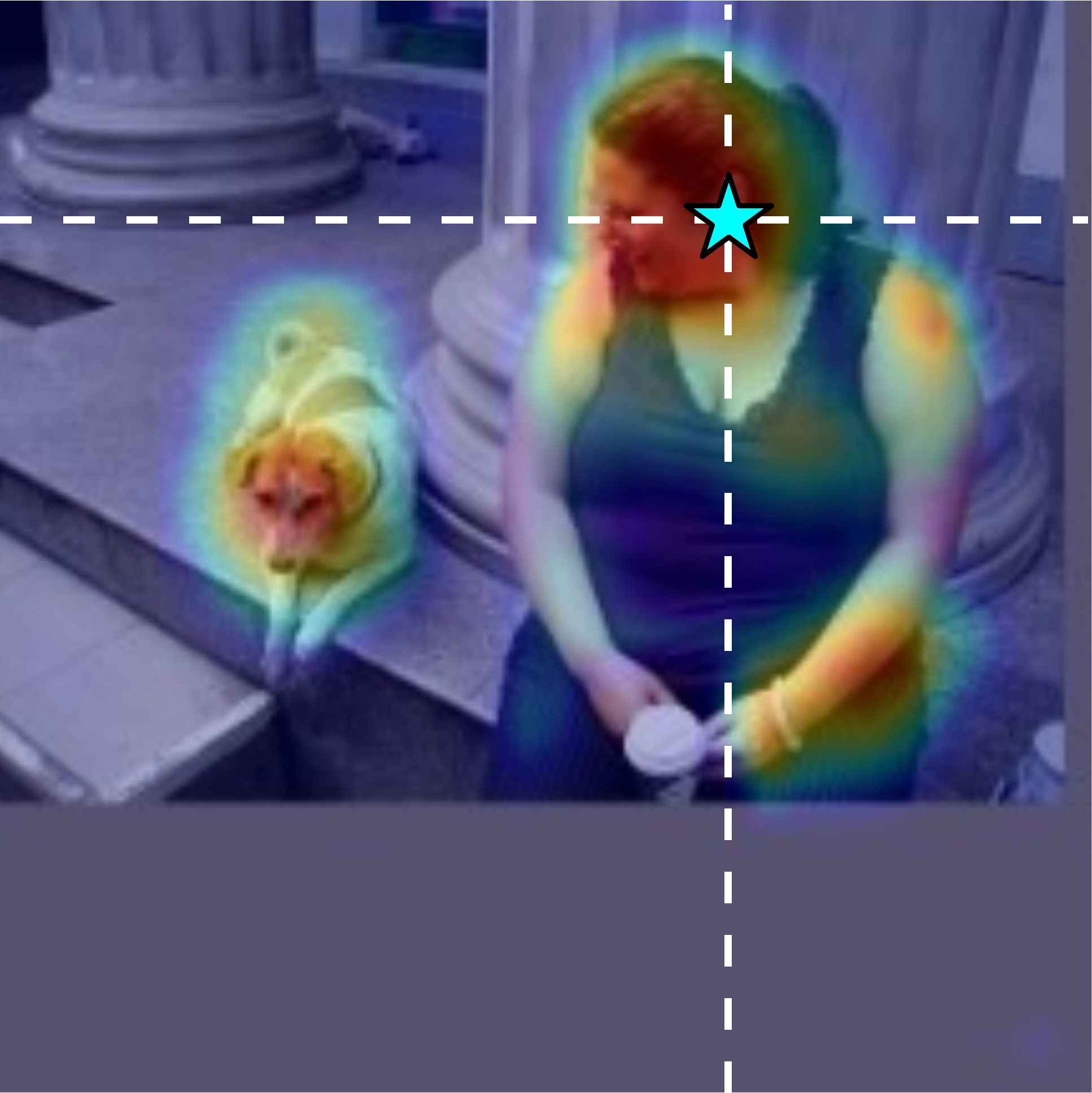}
         \caption{\texttt{Max}}
         \label{fig:Max}
    \end{subfigure}
    \begin{subfigure}[b]{0.24\linewidth}
        \centering
         \includegraphics[width=\textwidth]{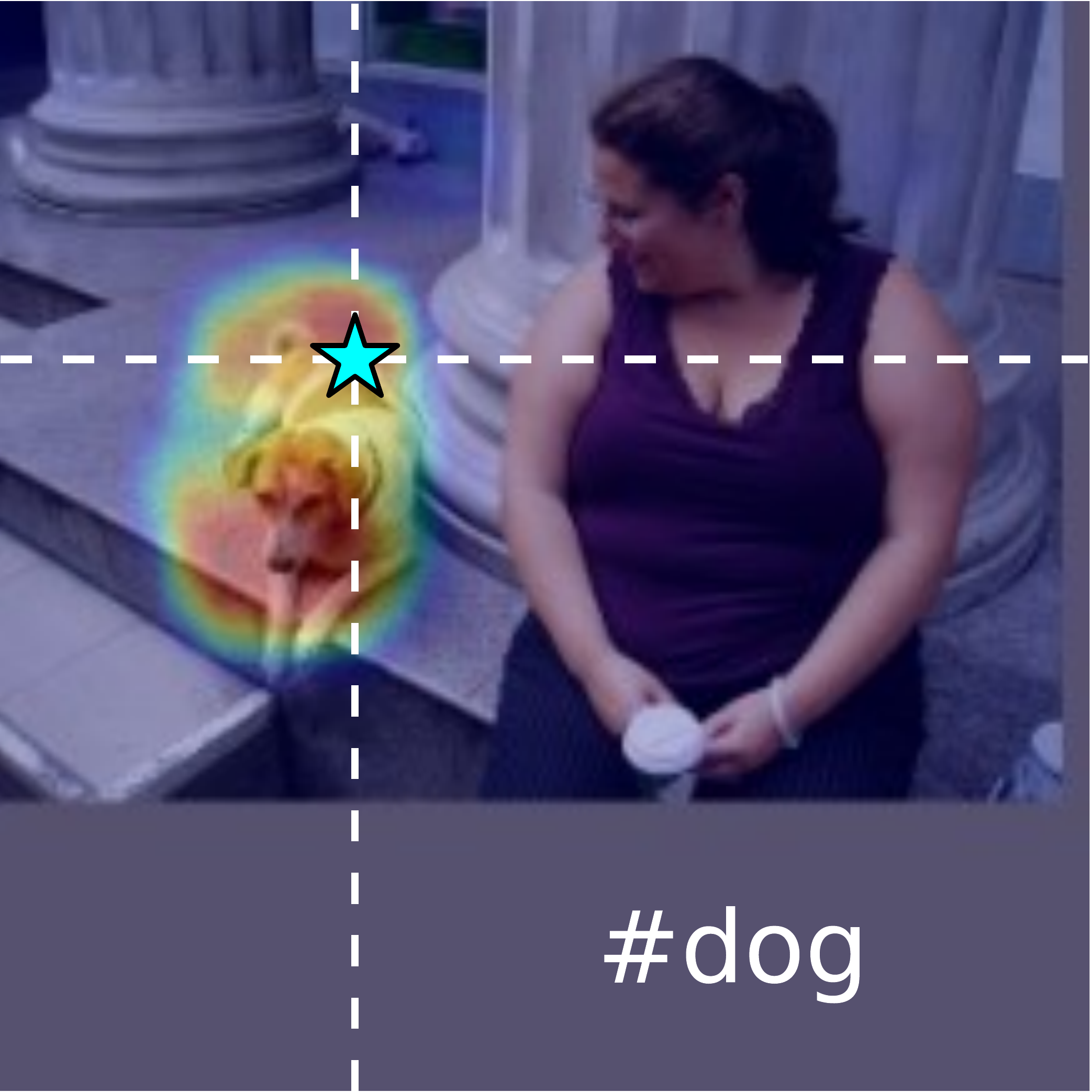}
         \caption{\texttt{cMax}}
         \label{fig:cMax}
    \end{subfigure}
    \begin{subfigure}[b]{0.24\linewidth}
        \centering
         \includegraphics[width=\textwidth]{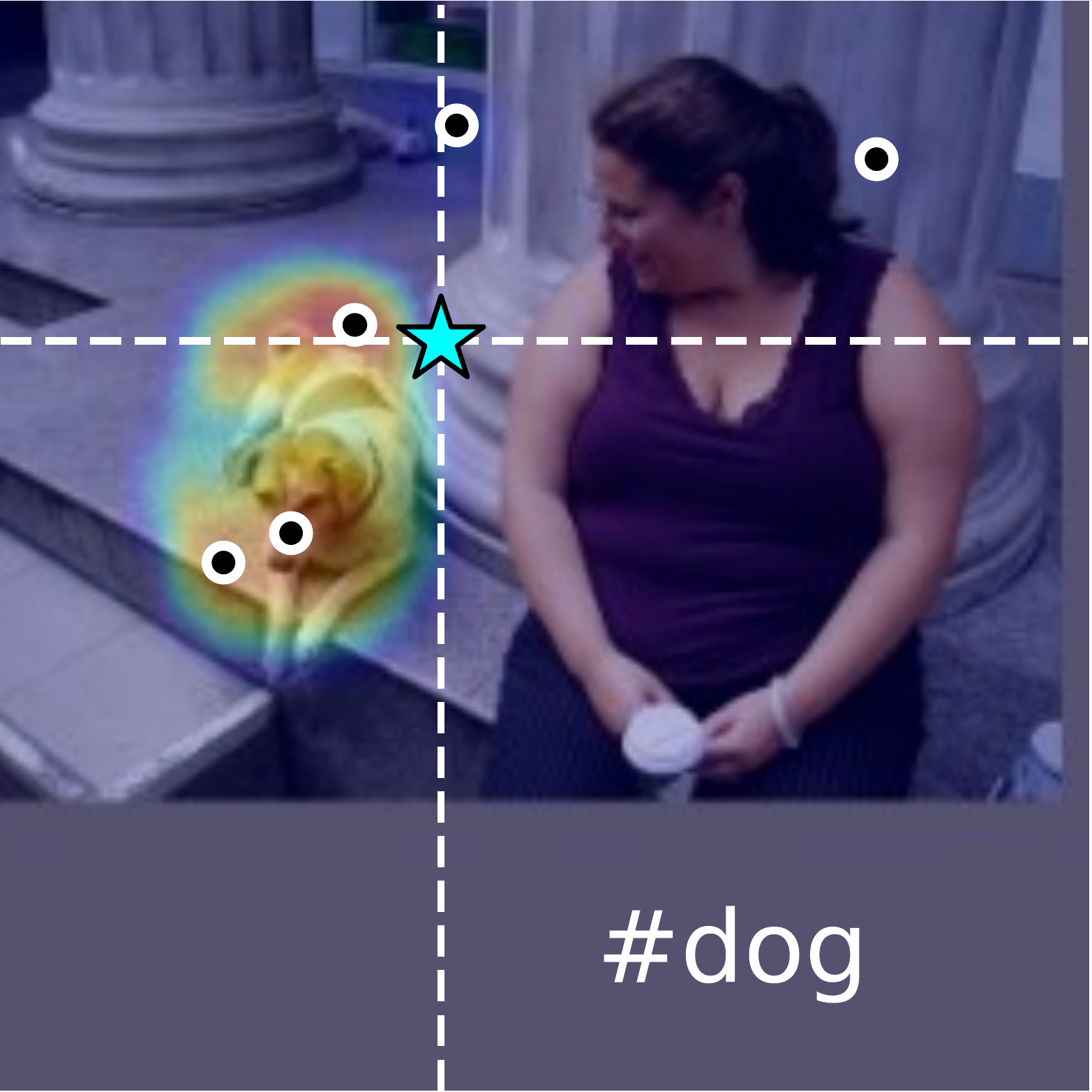}
         \caption{\texttt{cTopk}}
         \label{fig:ctopk}
    \end{subfigure}
    \begin{subfigure}[b]{0.24\linewidth}
        \centering
         \includegraphics[width=\textwidth]{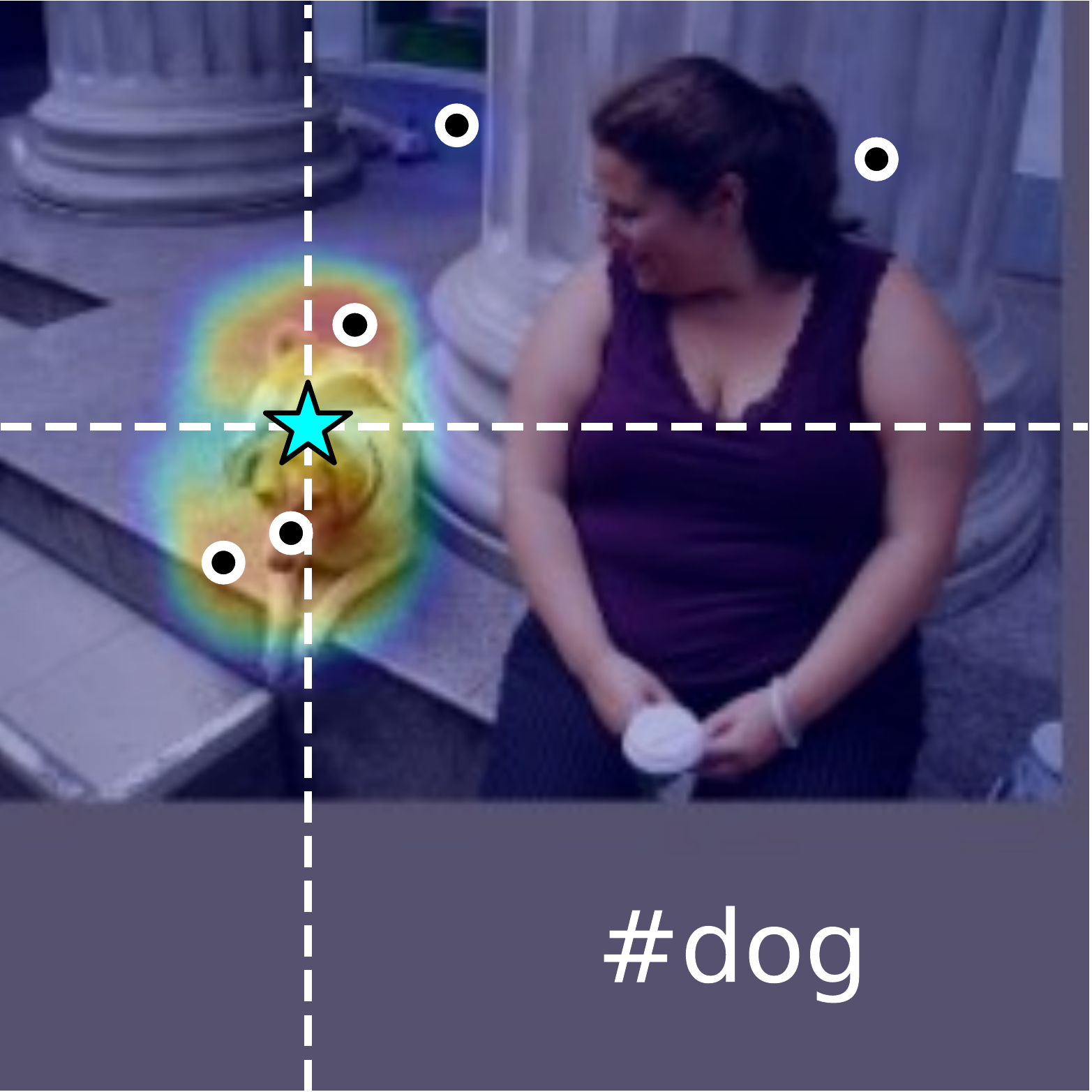}
         \caption{\texttt{cTopk-w}}
         \label{fig:ctopkw}
    \end{subfigure}\vspace{-8pt}
    \caption{Illustration of keypoint detection strategies. The black-white dots denote local maxima points, and the blue-black stars denote the calculated keypoints used for patch generating. The white dashed lines guide the patches generation.}
    \label{fig:kps}\vspace{-12pt}
\end{figure}

\section{Multi-label Representation Learning and Relation Learning}

\subsection{Object-Aware Self-supervision}
\label{sec:oass}
As stated in the above sections, our motivation is to impose effective self-supervision on objects under the MLL scenario. Hence we propose to generate object-Aware patches for the auxiliary task in SSL. Note that the advances of our self-supervision approach from the one proposed in \cite{puzzlecam} are not only in the patch generation strategy but also reflected in the modified so-called puzzle module, which will be elaborated below. 

We now explain the procedure of OASS. Given the input image $\bm{I}$ and the classification model $g(f(\bm{x};\bm{\theta});\bm{\phi})$, where $f(.)$ denotes the encoder part of the model parameterized by $\bm{\theta}$, and $g$ denotes the classification head. The $\bm{A} = f(\bm{x};\bm{\theta})$ be the output activation from the encoder $f$. The $\bm{A}_c = \mathrm{CAM}(\bm{A};g)\in\mathbb{R}^{C\times h\times w}$ represents the Class Attention Maps (CAMs) associated with feature maps $\bm{A}$, where $C$ is the number of the label.
Next, we cut the image into 4 patches, but instead of cutting them evenly as in \cite{puzzlecam}, we first calculate a keypoint $\bm{k}^\prime$ on the CAMs as follows:
\vspace{-6pt}
\begin{equation}\label{eq:keypoint}
\vspace{-6pt}
    \bm{k}^\prime = (k_x^\prime, k_y^\prime) = \arg \max \{\max{(\bm{A}_c^0, \bm{A}_c^1, ..., \bm{A}_c^{C-1})}\},
\end{equation}
which means $\bm{k}^\prime$ is the coordinate of the peak value on the merged CAM. We then obtain the corresponding location on the original image $\bm{k}=s\cdot \bm{k}^\prime$ where $s$ is the scaling factor of the encoder $f$ posed on the image(\emph{e.g. $s=16$} for ResNet-50). Then we cut the images into 4 patches.
Apart from the above key point detecting strategy (\texttt{Max}), we explore 3 additional strategies to potentially find more robust key points as below. The illustration of the below strategies is shown in \cref{fig:kps}.

\noindent\textbf{Channel-wise Max} (\texttt{cMax}) Instead of performing self-supervision on the merged CAM $\max{(\bm{A}_c^0, \bm{A}_c^1, ..., \bm{A}_c^{C-1})}$, we apply such a strategy to each channel of CAM $\bm{A}_c$ and generate $C$ reconstruction loss $\mathcal{L}_{re}^c$ and reconstructed classifcation loss $\mathcal{L}_{p-cls}^c$.

\noindent\textbf{Channel-wise Top-k} (\texttt{cTopk}) Similar to \texttt{cMax}, for each CAM channel, we select $k$ local maxima $\{\bm{p}_\kappa\}_{\kappa=1}^k$ on CAM with the top CAM values and calculate the keypoint as their geometric center: 
\vspace{-10pt}
\begin{equation}\label{eq:ctopk}
\vspace{-6pt}
    \bm{k}_c^\prime= \mathrm{round}\left(\frac{1}{k}\sum_\kappa^k {\bm{p}_\kappa^c}\right).\nonumber
\end{equation}

\noindent\textbf{Channel-wise weighted Top-k} (\texttt{cTopk-w}) Similar to \textbf{cTopk}, we calculate the geometric center by weighted averaging, where the weights are the corresponding normalized CAM responses: 
\vspace{-10pt}
\begin{equation}\label{eq:ctopkw}
\vspace{-6pt}
    \bm{k}_c^\prime= \mathrm{round}\left(\frac{1}{k}\sum_\kappa^k {\frac{\bm{A}_c(\bm{p}_\kappa^c)}{\sum_i^k{\bm{A}_c(\bm{p}_i^c)}}  \cdot \bm{p}_\kappa^c}\right).\nonumber
\end{equation} 

This will desirably split the object into each patch even when it doesn't fall on the center of the image. After generating the patches, we then have to resize them into unified sizes before entering the encoder again to get the patch-level feature maps hence all patches become the size of $(H / 2, W / 2)$. Next, we tile the generated patch-level feature maps similar to \cite{puzzlecam}, but we cannot tile them directly since the images patches are no longer the same size to begin with.
Hence we resize each patch feature map back to the size of its original image patch divided by $s$ before patch-level features tiling. 

\noindent\textbf{EMA for Patch-level Feature Extraction.} Inspired by recent knowledge distillation~\cite{tarvainen2017mean} and semi-supervised learning approaches, we extract patch-level features using the Exponential Moving Average~(EMA) of the encoder $f$ instead of sharing weight strategy in \cite{puzzlecam}, so that the patch-level features generated by a self-ensembled teacher is more stable and accurate.

\noindent\textbf{Loss Design.} 
We preserve the loss terms definitions from~\cite{puzzlecam}, including reconstruction loss $\mathcal{L}_{re}=||A^s-A^{re}||_1$ between original features $A^s$ and tiled ones $A^{re}$, and additional classifcation loss of logits from $A^{re}$: $\mathcal{L}_{p-cls} = l_{cls}(\hat{Y}^{re}, Y)$. 
The final total training loss for latter 3 channel-wise strategies is: 
\vspace{-8pt}
\begin{equation}\label{eq:loss}
\vspace{-6pt}
\mathcal{L} = \mathcal{L}_{cls} + \frac{\sum_{c}^C{\mathbb{I}(\bm{y}_c=1)(\alpha_{p}\mathcal{L}_{p-cls}^c + \alpha_{re}\mathcal{L}_{re}^c)}}{|\{\mathbb{I}(\bm{y}_c=1)\mid 1\le c\le C\}|},\nonumber
\end{equation}
where $\alpha_{re},\alpha_p$ is the loss weights for $\L_{re},\L_{p-cls}$ respectively, and $\bm{y}_c$ is the classification label for class $c$.

To this end, the proposed Object-Aware Self-supervision forces the encoder feature maps to contain as much object-related information to facilitate the classification on the reconstructed feature maps, which in turn helps the encoder to extract more discriminative features.

\subsection{Instance-level MLL with Class-specific Instance~(CSI)}

\noindent\textbf{Class-specific Instance Generation Module} As mentioned in \cref{sec:oass}, the encoder trained on OASS can produce robust features for multi-object. Such features can be easily reorganized by an additional sub-network, categorizing them into different class-specific semantics. Therefore, as the Class-specific instance generation module in Fig.~\ref{fig:pipeline} shows, we mask the features $f$ learned by OASS with the CAMs to extract such class-specific semantics on top of the encoder output and transform the encoder features into class-specific instances $\{f_1, f_2, ..., f_C\}$. 


\noindent\textbf{Instance-level Multi-label Learning}
Next, we build a MLL classification head on top of class-specific instances (CSI). We perform pixel-wise classification with a $1\times1$ Conv layer with $1$ output channel for each $C$ CSI feature map. Then we concatenate these $C$ logits maps along channels before further performing an additional $1\times1$ Conv. Finally, we perform a global average pooling to obtain image-level MLL logits.

\section{Experiments}

\subsection{Implementation Details}

We validate our proposed framework on Pascal VOC2012 dataset~\cite{everingham2010pascal} to align with our major baseline PuzzleCAM~\cite{puzzlecam}. For fairness, we train our models on only \texttt{train} split same as ~\cite{puzzlecam} throughout the experiments and evaluate the models on \texttt{val} split (10,582 training and 1,449 validation images). Same with \cite{puzzlecam}, firstly, all images are randomly resized between $320$ and $640$ followed by horizontal flipping. All samples are padded to $512\times 512$ for training and inference. We set target loss weights $\alpha_{re}=\alpha_p=\frac{1}{15}$ for $\mathcal{L}_{re}$ and $\mathcal{L}_{p-cls}$ respectively, and ramped up both $\alpha_{re}$ and $\alpha_{p}$ linearly to $100$ epochs. We adopt ResNet-50 pre-trained on ImageNet as the backbone feature extractor throughout and optimize the model with Adam optimizer. We conduct all experiments on Nvidia's A100 GPUs with 48G VRAM.

\subsection{Comparison with existing methods}

We conducted extensive experiments on VOC2012 benchmark dataset to demonstrate our advances in Multi-label Classification.  

\begin{figure}[t]
    \centering
    \includegraphics[width=.75\linewidth]{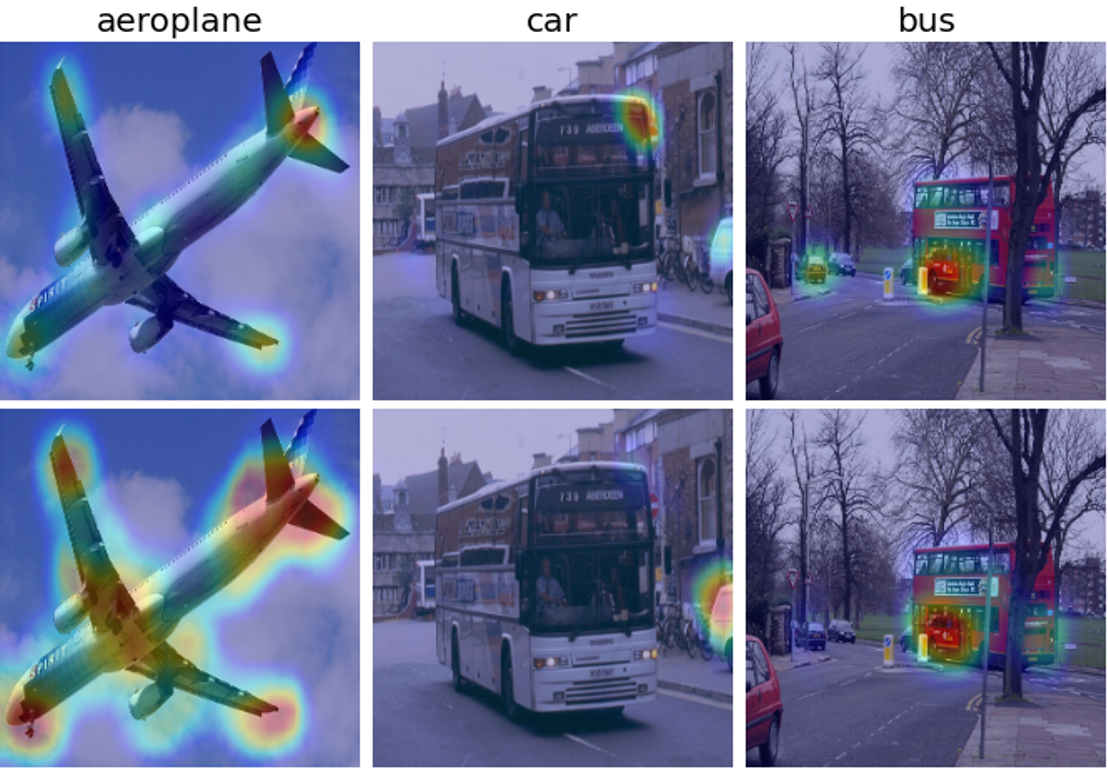}
    \caption{Visualizations of the CAMs generated on \texttt{val} split. The first row is generated by PuzzleCAM method, and the second row is generated by our method (\texttt{cTopk-w} w/o CSI).}\vspace{-8pt}
    \label{fig:vis_cam}
\end{figure}


\begin{table}[ht]
    \centering
    \begin{tabular}{c|c|c}\hline
        Method & Architecture & mAP \\\hline
        SSGRL~\cite{chen2019learning} & ResNet-101 & $91.9$ \\
        SSGRL~\cite{chen2019learning} & ResNet-50 & $91.4$  \\
        MCAR~\cite{MCAR_TIP_2021} & ResNet-50 & $90.04$ \\
        PuzzleCam~\cite{puzzlecam} & ResNet-101 & $92.60$ \\
        PuzzleCam~\cite{puzzlecam} & ResNet-50 & $89.3$ \\\hline
        Ours (w/ gt-bbox) & ResNet-50 & $92.62$ \\\hline
        Ours (\texttt{Max}) & ResNet-50 & $\mathbf{92.45}$ \\
        Ours (\texttt{cTopk-w}) & ResNet-50 & $92.20$ \\\hline
    \end{tabular}
    \caption{Comparison of classification mAP on VOC2012 \texttt{val} split with state-of-the-art methods.}\vspace{-8pt}
    \label{tab:com}
\end{table}
\begin{table}[!h]
    \centering
    \begin{tabular}{c|c}\hline
        Method & mAP \\\hline
        PuzzleCAM\cite{puzzlecam} & $81.37$ \\\hline
        Ours (\texttt{Max}) w/ CSI & $\mathbf{84.27}$ \\
        Ours (\texttt{cTopk-w}) w/o CSI & $\mathbf{88.43}$ \\\hline
    \end{tabular}
    \caption{mAP on \texttt{test} split compared to baseline.}\vspace{-12pt}
    \label{tab:test}
\end{table}

As shown in \cref{tab:com}, our approach with \texttt{Max} achieves $92.45$ mAP scores on \texttt{val} split, which greatly outperform our strong baseline PuzzleCAM~\cite{puzzlecam} for both splits. 
Ours (\texttt{gt-bbox}) in \cref{tab:com} and \cref{tab:abl_kps} represents the upper bound of our method, where the keypoints are derived from the auxillary bounding box annotations from the VOC2012 dataset, which are calculated as the center of the corresponding bounding boxes.
On VOC2012 \texttt{test} split, as shown in \cref{tab:test}, our method exceeds our sole baseline by a large margin.
Apart from quantitative results, \cref{fig:vis_cam} displays generated CAMs, showing that our method can not only promote the recall of more detailed contextual instance features but also suppress the features from non-related semantics. Such local contextual features are not only naturally beneficial to the classification but also provide abundant material to generate class-specific instances. The baseline~\cite{puzzlecam} fails to extract good enough local features, thus underperforming our approach.

\begin{figure}[t]
    \centering
    \includegraphics[width=0.82\linewidth]{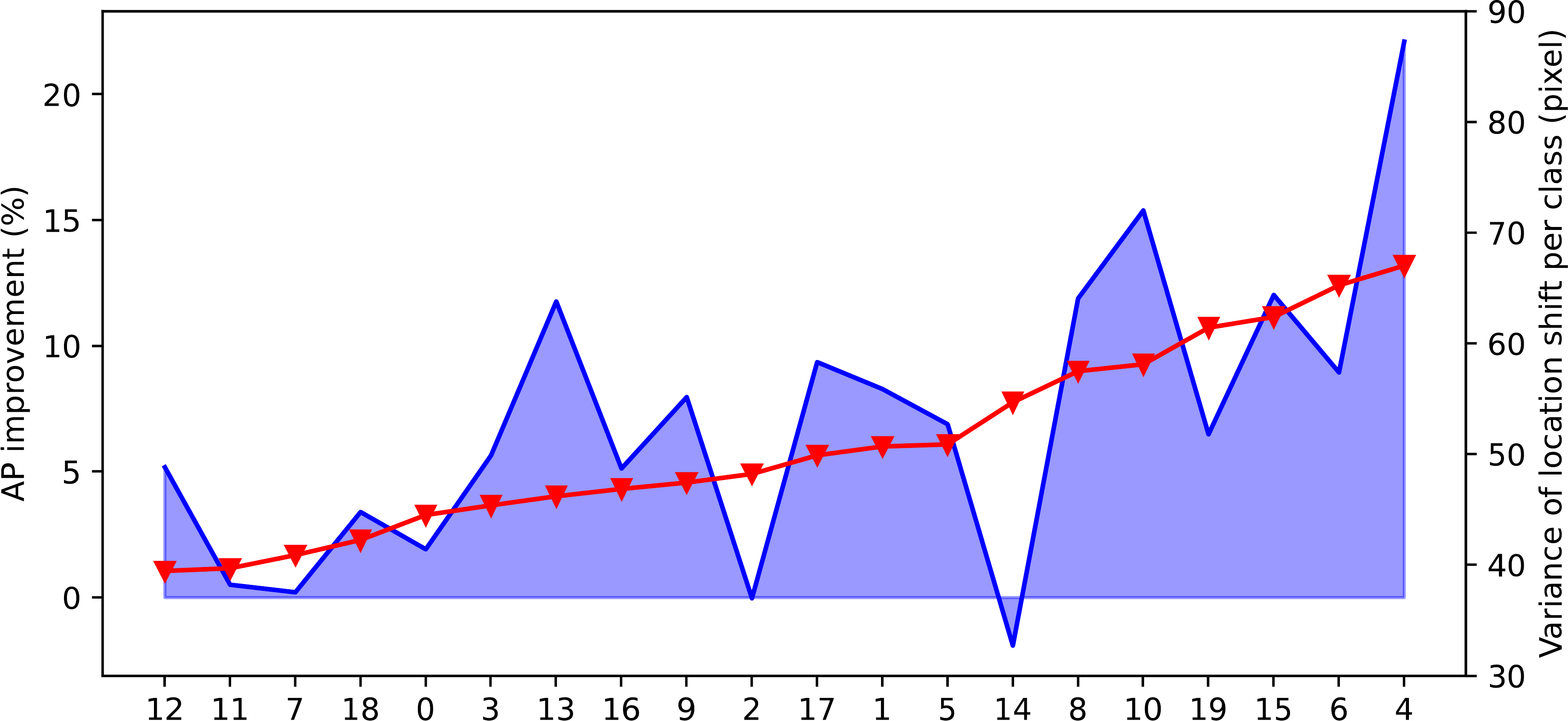}
    \caption{Per-class AP improvements (\textcolor{blue}{blue} area) and corresponding variances of object location shifts (\textcolor{red}{red} curve). X-axis denotes class index.}\vspace{-8pt}
    \label{fig:improve and shift}
\end{figure}

\subsection{Ablation Studies}

\begin{table}[!thb]
    \centering
    \begin{tabular}{c|c|c}\hline
        Method & w/o CSI & w/ CSI \\\hline
        Ours (w/ gt-bbox) & $92.65$ &$92.62$\\\hline
        Ours (\texttt{Max}) & $\mathbf{92.38}$ & $\mathbf{92.45}$ \\
        Ours (\texttt{cMax}) & $91.58$ & $92.08$ \\
        Ours (\texttt{cTopk}) & $92.21$ & $92.05$ \\
        Ours (\texttt{cTopk-w}) & $92.02$ & $92.20$ \\\hline 
    \end{tabular}
    \caption{Ablation studies on keypoint detection strategies.}
    \label{tab:abl_kps}\vspace{-8pt}
\end{table}

\noindent\textbf{Effect of different keypoint detection strategies.}
First, we explore the effect of different keypoint detection strategies for Object-Aware Self-supervision. We first remove the CSI module and test different strategies in OASS module. As shown in \cref{tab:abl_kps}, among different proposed keypoint detecting strategies, \texttt{Max} performed the best on \texttt{val} split, while \cref{tab:test} shows that \texttt{cTopk-w} generalizes the best on \texttt{test} split. As can be seen from \cref{fig:kps}, this is because \texttt{cTopk-w} has less chance to be misled by low-quality global maxima on CAMs by considering other local maxima as in the case of \texttt{Max}. 

\cref{fig:improve and shift} presents the improvements of per-class AP scores on our method (\texttt{cTopk-w} w/o CSI) against baseline~\cite{puzzlecam} on \texttt{test} 1. Compared with the variances of the object location shifts statistics of different classes (\textcolor{red}{red} curve) derived from the statistics in \cref{fig:stats_box}, we can easily observe that they roughly follow the same trend. In other words, the larger the location of the object of certain class varies in images, the more likely the precision performance (AP) of this class benefits from our method, supporting the effectiveness of our proposed Object-Aware Self-supervision.

\noindent\textbf{Effect of Class-specific Instance.} Comparing w/o CSI and w/ CSI methods for both \texttt{cTopk-w} and \texttt{Max} strategies in \cref{tab:abl_kps}, w/ CSI method constantly performs better than w/o CSI ones, proving the effectiveness of CSI.

\section{Conclusions}

In this paper, we proposed Object-Aware Self-supervision to facilitate local contextual representation learning for multi-label circumstances. 
To facilitate instance-level MLL, we proposed to generate Class-specific Instance harnessing the self-supervision outcome.
Extensive experiments and ablation studies validated the proposed approach for multi-label learning. We plan to apply a stronger CNN backbone and adapt our method to weakly supervised segmentation in the future.

\clearpage
\bibliographystyle{ieee_fullname}
\bibliography{ref}

\end{document}